\documentclass[sigconf,screen]{acmart}
\setcopyright{acmcopyright}
\acmYear{2020}
\copyrightyear{2020}
\acmConference[RL+SE\&PL'20, Co-located with ESEC/FSE]{Proceedings of the 1st ACM SIGSOFT International Workshop on Representation Learning for Software Engineering and Program Languages}{2020}{Virtual, USA}
\acmBooktitle{Proceedings of the 1st ACM SIGSOFT International Workshop on Representation Learning for Software Engineering and Program Languages (RL+SE\&PL'20, Co-located with ESEC/FSE), 2020, Virtual, USA}

\usepackage{fancybox}
\usepackage{graphicx}
\usepackage{float}
\usepackage{pbox}
\usepackage{listings}
\usepackage{amssymb}
\usepackage{url} 
\usepackage{hyperref}
\usepackage{adjustbox}
\usepackage{multirow}
\usepackage{xspace}
\usepackage{comment}
\usepackage{xcolor}
\usepackage{microtype}
\usepackage{balance}

\title{
    Towards Demystifying Dimensions of Source Code Embeddings
}

\author{Md Rafiqul Islam Rabin}
\affiliation{%
   \institution{University of Houston}
}
\email{mrabin@central.uh.edu}

\author{Arjun Mukherjee}
\affiliation{%
   \institution{University of Houston}
}
\email{amukher6@central.uh.edu}

\author{Omprakash Gnawali}
\affiliation{%
   \institution{University of Houston}
}
\email{odgnawal@central.uh.edu}

\author{Mohammad Amin Alipour}
\affiliation{%
   \institution{University of Houston}
   }
\email{maalipou@central.uh.edu}

\ccsdesc[500]{Computing methodologies~Learning latent representations}
\ccsdesc[500]{Software and its engineering~General programming languages}

\keywords{Code Representation, Code Embeddings, Models of Code}

\begin{document}
\newcommand{\JS}{\textsc{Java-Small}\xspace}
\newcommand{\JM}{\textsc{Java-Med}\xspace}
\newcommand{\JL}{\textsc{Java-Large}\xspace}

\newcommand{\Fix}[1]{\textbf{\textcolor{red}{Fix/TODO}: #1}}
\newcommand{\Ans}[1]{\textbf{\textcolor{blue}{Answer}: #1}}
\newcommand{\Part}[1]{\noindent\textbf{#1}}
\newcommand{\fsize}[2]{{\fontsize{#1}{0}\selectfont#2}}

\newcommand{\eg}{\textit{e.g.}\xspace}
\newcommand{\ie}{\textit{i.e.}\xspace}
\newcommand{\etal}{\textit{et al.}\xspace}

\newcounter{observation}
\newcommand{\observation}[1]{\refstepcounter{observation}
        \begin{center}
        \Ovalbox{
        \begin{minipage}{0.9\columnwidth}
            \textbf{Observation \arabic{observation}:} #1
        \end{minipage}
        }
        \end{center}
}

\newcommand{\TopTen}{\textsc{Top-Ten}\xspace}
\newcommand{\Fone}{\textsc{$F_1$-Score}\xspace}
\newcommand{\SVMHC}{\textsc{SVM-Handcrafted}\xspace}
\newcommand{\SVMNN}{\textsc{SVM-code2vec}\xspace}
\newcommand{\GRUCH}{\textsc{GRU-CharSeq}\xspace}
\newcommand{\GRUTK}{\textsc{GRU-TokenSeq}\xspace}
\newcommand{\HC}{\textsc{Handcrafted}\xspace}
\newcommand{\CtV}{\textsc{code2vec}\xspace}
\newcommand{\SVML}{$SVM^{light}$\xspace}
\newcommand{\SVMHCS}{\textsc{SVM-Handcrafted(Norm,Binary,Mix)}\xspace}

\newcommand{\HCB}{\textsc{HC(Binary)}\xspace}
\newcommand{\HCN}{\textsc{HC(Norm)}\xspace}
\newcommand{\CX}{\textsc{CX(Norm)}\xspace}
\newcommand{\HCBCX}{\textsc{HC(Binary)+CX(Norm)}\xspace}
\newcommand{\HCNCX}{\textsc{HC(Norm)+CX(Norm)}\xspace}
\newcommand{\HCM}{\textsc{HC(Mix)}\xspace}
\newcommand{\BGCH}{\textsc{CharSeq}\xspace}
\newcommand{\BGTK}{\textsc{TokenSeq}\xspace}

\newcommand{\NMF}{33\xspace}
\newcommand{\NCF}{14\xspace}
\newcommand{\NTF}{47\xspace}
\newcommand{\NCH}{94\xspace}
\newcommand{\NTK}{108106\xspace}

\newcommand{\crossmark}{$\times$}


\begin{abstract}
Source code representations are key in applying machine learning techniques for processing and analyzing programs. A popular approach in representing source code is \textit{neural source code embeddings} that represents programs with high-dimensional vectors computed by training deep neural networks on a large volume of programs. Although successful, there is little known about the contents of these vectors and their characteristics.

In this paper, we present our preliminary results towards better understanding the contents of code2vec neural source code embeddings. In particular, in a small case study, we use the code2vec embeddings to create binary SVM classifiers and compare their performance with the handcrafted features. Our results suggest that the handcrafted features can perform very close to the highly-dimensional code2vec embeddings, and the information gains are more evenly distributed in the code2vec embeddings compared to the handcrafted features. We also find that the code2vec embeddings are more resilient to the removal of dimensions with low information gains than the handcrafted features. We hope our results serve a stepping stone toward principled analysis and evaluation of these code representations.
\end{abstract}

\maketitle
\renewcommand{\shortauthors}{MRI Rabin, A Mukherjee, O Gnawali, MA Alipour}

\section{Introduction}
\label{sec:introduction}
The availability of a large number of mature source code repositories has fueled the growth of ``Big Code'' that attempts to devise data-driven approaches in the analysis and reasoning of the programs~\cite{BigCodeSurvey, chen2019literature} by discovering and utilizing commonalities within software artifacts.
Such approaches have enabled a host of exciting applications \eg, prediction of data types in dynamically typed languages~\cite{vincent:type}, detection of the variable naming issues~\cite{allamanis2017learning}, or repair of software defects~\cite{dinella2020hoppity}. 

Deep neural networks have accelerated innovations in Big Code and have greatly enhanced the performance of prior traditional approaches. 
The performance of deep neural networks in cognitive tasks such as method name prediction~\cite{allamanis2016convolutional} or variable naming~\cite{allamanis2017learning} has reached or exceeded the performance of other data-driven approaches. 
The performance of neural networks has encouraged researchers to increasingly adopt the neural networks in processing source code.

Source code representation is the cornerstone of using neural networks in processing programs. 
Numerous work on devising representations for code in certain tasks \cite{BigCodeSurvey, chen2019literature}.
In such representations, the code is represented by a vector of numbers, called embeddings, resulted from training on millions of lines of source code or program traces.
The current state of practice in devising such representations includes decisions about the length of code embeddings, code features included in learning, etc.
The current approach is highly empirical and tedious; moreover, the analysis and evaluation of  the source embeddings are nontrivial.

While there are an increasing number of work on the interpretation and analysis of neural networks for source code, \eg, \cite{Nghi2019AutoFocus}, \cite{rabin2019tnpa}, \cite{Kang2019Generalizability}, and \cite{rabin2020generalizability}, to the best of our knowledge there is no work to look at the internal of source code embeddings.
In addition to facilitating the interpretation of the behavior of neural models, understanding the source code embeddings would enable researchers and practitioners to optimize neural models, and potential can provide methodologies to objectively compare different representations.

In this work, we report our initial attempts for demystifying the dimensions of source code embeddings, which is aimed at a better understanding of the embedding vectors by analyzing their values and comparing them with understandable features.
In particular, we report the results of our preliminary analysis of code2vec embeddings \cite{alon2019code2vec}, a popular code representation for method name prediction task.
More specifically, we use the code2vec embeddings to build SVM models and compare them with SVM models trained on the handcrafted features. We analyze the statistical characteristics of the dimensions in the embeddings.

Our results suggest that the handcrafted features can perform very close to the highly-dimensional code2vec embeddings, and the information gains are more evenly distributed in the code2vec embeddings compared to the handcrafted features.
We also find that the code2vec embeddings are more resilient to the removal of dimensions with low information gains than the handcrafted features.

\Part{Contributions.}
This paper makes the following contributions.
\begin{itemize}
    \item It provides an in-depth analysis of dimensions in code2vec source code embeddings in a small number of methods.
    \item It compares the performance of handcrafted features with naive representations and code2vec embeddings.
\end{itemize}

\section{Background}
\label{sec:background}

The code2vec \cite{alon2019code2vec} source code representation uses bags of paths in the abstract syntax tree (AST) of programs to represent programs. 
The model encodes the AST path between leaf nodes and uses an attention mechanism to compute a learned weighted average of the path vectors in order to produce a single code vector of $384$ dimensions for each program. 

\begin{figure}[H]
    \centering
    \includegraphics[width=\columnwidth]{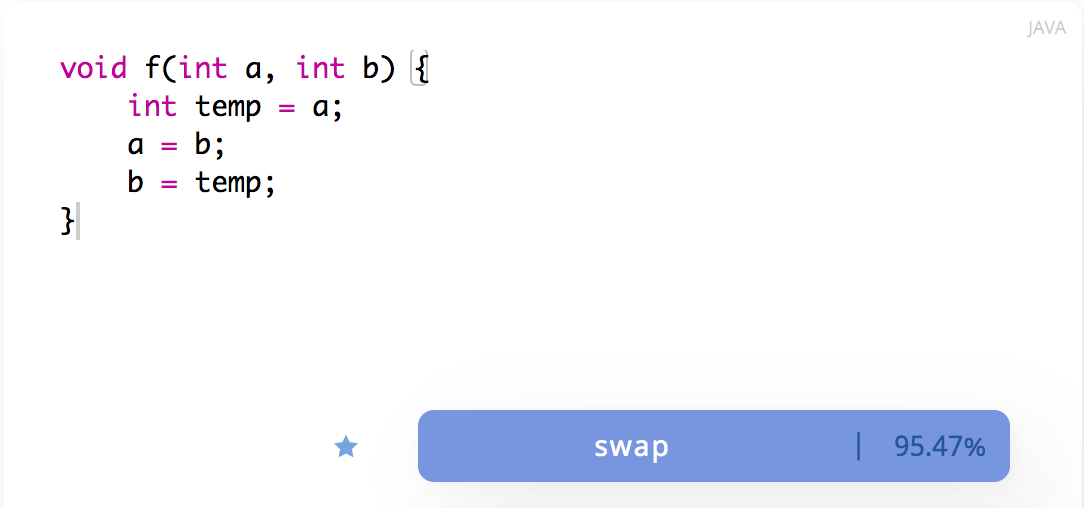}
    \caption{An example of method name prediction by code2vec\cite{alon2019code2vec}.}
    \label{fig:example}
\end{figure}

The code2vec \cite{alon2019code2vec} was initially introduced to predict the name of method \cite{allamanis2016convolutional}, given the method's body.
Figure~\ref{fig:example} depicts an example of this task wherein a neural model based on code2vec correctly predicts the name of the method in the Figure as \texttt{swap}.


\section{Methodology}
\label{sec:methodology}

\begin{figure*}[!t]
    \centering
    \includegraphics[scale=.48]{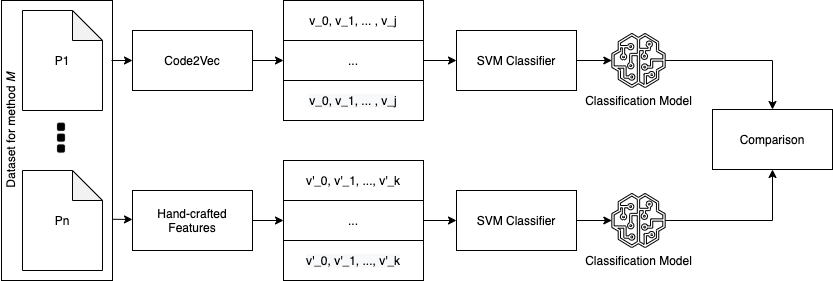}
    \caption{Workflow in this study.}
    \label{fig:workflow}
\end{figure*}

To evaluate the code2vec code representation we follow the workflow in Figure~\ref{fig:workflow}. 
We first select a few methods in which we are interested in the analysis of their representations. 
We then manually select features that best can predict their names. 
Next, we create binary classifiers for predicting the name of those methods with code2vec embeddings and handcrafted features.   
Finally, we evaluate and compare the performance of the trained classification models.
In the rest of this section, we will describe dataset and selection of methods, feature extraction, classifier creation, baseline classifiers, and evaluation metrics.

\subsection{Dataset and Method Selection}

\Part{\TopTen dataset}.
We use the \JL dataset~\cite{alon2019code2seq} that contains 9K Java projects in the training set, 200 Java projects in the validation set, and 300 Java projects in the test set that were collected from GitHub. 
Overall, it contains about 16M methods where almost 3.5M methods have a unique name.
We chose ten most-frequent method names in the \JL dataset and the corresponding method bodies to create a new dataset, \TopTen, for further analysis. 

The reason for restricting our analysis to these methods is twofold. First, the sheer number of method names in \JL prohibits a scalable manual inspection and analysis for all methods. Second, the distribution of method names in \JL conforms to the power-law; that is, relatively few method names appear frequently in the dataset while the rest of method names appear rarely in the dataset. Therefore, the performance of any classifier on \JL heavily relies on its performance on the few frequent method names. Column ``Name of Method'' in Table \ref{table:top10_features} lists the names of ten most-frequent methods that we chose for our analysis.

\Part{Deduplication of the \TopTen dataset.}
As noted by \cite{Miltos2019Duplication}, the \JL dataset suffers from duplicate methods that can inflate the results of the prediction. 
We removed duplicate methods in the dataset following the steps outlined in \cite{Miltos2019Duplication} and used the same parameters for deduplication thresholds: key-jaccard-threshold, $t_0 = 0.8$ and jaccard-threshold, $t_1 = 0.7$.

\Part{Dataset for each \TopTen method.}
For each method \textit{M} in \TopTen, we create a training set that constitutes from $1000$ randomly selected positive examples (methods with name \textit{M}), and $1000$ randomly selected negative examples (any method but \textit{M}) from the deduplicated \TopTen training set.
For the validation set and test set, we select all the positive examples and the same number of randomly selected negative examples from the deduplicated \TopTen validation set and test set, respectively.
Table \ref{table:top10_db} shows the size of the dataset for each \TopTen method.

\subsection{Extracting Handcrafted Features}

\begin{table} 
    \begin{center}
        \def\arraystretch{1.1}
        \caption{\TopTen method name and feature list.}
        \label{table:top10_features}
        \resizebox{\columnwidth}{!}{%
        \begin{tabular}{ |c|l| }
            \hline
            \textbf{Name of Method} & \textbf{Feature List} \\ \hline
            \hline
            
            \multirow{1}{*}{equals}
            & Instance, Boolean, equals, This
            \\ \hline
            
            \multirow{1}{*}{main}
            & Println, String \\ \hline
            
            \multirow{1}{*}{setUp}
            & Super, setup, New, build, add \\ \hline
            
            \multirow{1}{*}{onCreate}
            & Bundle, onCreate, setContentView, R \\ \hline
            
            \multirow{1}{*}{toString}
            & toString, format, StringBuilder, append, + \\ \hline
            
            \multirow{1}{*}{run}
            & Handler, error, message \\ \hline
            
            \multirow{1}{*}{hashCode}
            & hashCode, TernaryOperator \\ \hline
            
            \multirow{1}{*}{init}
            & init, set, create \\ \hline
            
            \multirow{1}{*}{execute}
            & CommandLine, execute, response \\ \hline
            
            \multirow{1}{*}{get}
            & Return, get \\ \hline

        \end{tabular}%
        }
    \end{center}
\end{table}

\begin{table} 
    \begin{center}
        \def\arraystretch{1.1}
        \caption{Additional code complexity features.}
        \label{table:complexity_features}
        \resizebox{\columnwidth}{!}{%
        \begin{tabular}{| l |}
            \hline
            \texttt{LOC, Block, Basic Block, Parameter, Local Variable,} \\
            \texttt{Global Variable, Loop, Jump, Decision, Condition,} \\
            \texttt{Instance, Function, TryCatch, Thread} \\
            \hline
        \end{tabular}%
        }
    \end{center}
\end{table}

\begin{table} 
    \begin{center}
        \def\arraystretch{1.1}
        \caption{Size of the dataset for each \TopTen method.}
        \label{table:top10_db}
        \resizebox{0.98\columnwidth}{!}{%
        \begin{tabular}{ |c|c|c|c| }
            \hline
            \textbf{Name of Method} & \textbf{\#Training} & \textbf{\#Validation} & \textbf{\#Test} \\ \hline
            \hline
            
            equals   & 2000 & 1212 & 1778 \\ \hline
            main     & 2000 & 1220 & 2032 \\ \hline
            setUp    & 2000 & 1220 & 1424 \\ \hline
            onCreate & 2000 & 1876 & 1484 \\ \hline
            toString & 2000 & \phantom{0}586 & 1278 \\ \hline
            run      & 2000 & \phantom{0}876 & 1558 \\ \hline
            hashCode & 2000 & \phantom{0}534 & \phantom{0}770 \\ \hline
            init     & 2000 & \phantom{0}892 & 2504 \\ \hline
            execute  & 2000 & \phantom{0}498 & \phantom{0}702 \\ \hline
            get      & 2000 & \phantom{0}780 & \phantom{0}670 \\ \hline

        \end{tabular}%
        }
    \end{center}
\end{table}

\Part{Method-only features}.
For each method, two authors do their best effort to draw discriminant features by inspecting the training dataset.
Table \ref{table:top10_features} shows the handcrafted features for each method, in total, \NMF features for ten methods.

\Part{Code complexity features}.
An important metric of interest might be adding code complexity features. Similar methods may have certain patterns such as the number of lines, variables, or conditions. 
Therefore, we further extend the handcrafted features with an additional \NCF code complexity features shown in Table \ref{table:complexity_features}.
Thus, the handcrafted features become a union of \NTF features including the code complexity features.
Note that we only focus on the simpler code complexity features shown in Table \ref{table:complexity_features} as our study is limited to the methods, and thus the class level or project level code complexity metrics do not apply to our study.

\Part{Feature Extraction}.
We use the \texttt{JavaParser}~\cite{smith2017javaparser} tool to parse the methods in the dataset and extract the handcrafted features. 
We consider the \NMF handcrafted features of methods (Table \ref{table:top10_features}) as (a) binary vectors, and (b) numeric vectors. 
For the binary vectors, we use 1 and 0 that denote the presence or absence of individual features in the method, respectively. 
For the numeric vectors, we count the number of occurrences of features in the program, and in the end, we normalize them using StandardScaler \cite{scikit-learn} to map the distribution of values to a mean value of $0$ and a standard deviation of $1$. 
The \NCF complexity features (Table \ref{table:complexity_features}) are always considered as numeric values.

\subsection{Classification Models} 
\Part{Support Vector Machines}. 
Support Vector Machines (SVM) are one of the most popular traditional supervised learning algorithms that can be used for classification and regression on linear and non-linear data \cite{cortes1995support,hsu2003practical,ben2010guide}. SVM uses the concept of linear discriminant and maximum margin to classify between classes. Given the labeled training data points, SVM learns a decision boundary to separate the positive points from the negative points. The decision boundary is also known as the maximum margin separating hyperplane that maximizes the distance to the nearest data points of each class. The decision boundary can be a straight line classifying linear data in a two-dimensional space (i.e. linear SVM using linear kernel) or can be a hyperplane classifying non-linear data by mapping into a higher-dimensional space (i.e. non-linear SVM using RBF kernel).

\Part{Classifiers}.
For each method \textit{M}, we  create two SVM classification models: \SVMHC and \SVMNN. 
\SVMNN uses the code2vec embeddings of the programs in training the SVM model,
which is a single fixed-length code embedding (384 dimensions) that represents the source code as continuous distributed vectors for predicting method names. 
\SVMHC uses the vector of the handcrafted features (\NMF dimensions without complexity features, and \NTF dimensions with complexity features) to train an SVM model. 

\Part{Training}
We use the $SVM^{light}$ \footnote{\url{http://svmlight.joachims.org/}}, an implementation of Support Vector Machines (SVMs) in C \cite{joachims1999svm}, to train the classification models in the experiments. 

Since the performance of SVM depends on its hyper-parameters, 
we run the grid search algorithm \cite{bard1982grid} for hyper-parameter optimization. 
We train SVMs with tuned parameters on handcrafted features and code2vec embeddings for each method name.

\subsection{Naive Sequence-based Neural Baselines}
We also create two sequence-based baselines to compare our handcrafted features: (a) \BGCH where the program is represented by a sequence of characters in the program, and (b) \BGTK where a sequence of tokens in the program represent the program.

\Part{\BGCH}. For character-based representation, we first remove comments from the body of the method and save the body as a plain string. Then we create a list of ASCII \footnote{\url{https://en.wikipedia.org/wiki/ASCII} (character code 0-127 in ASCII-table)} characters by filtering out all non-ASCII characters from the string of body. After that, we create a character-based vocabulary with the unique ASCII characters found in the training+validation set of the \TopTen dataset (the character-based vocabulary stores \NCH unique ASCII characters). Finally, we encode the method body by representing each character with its index in the character-based vocabulary.

\Part{\BGTK}. For token-based representation, we modify the JavaTokenizer tool \cite{Miltos2019Duplication} to get the sequence of Java tokens from the body of the method. After that, we create a token-based vocabulary with the unique tokens found in the training+validation set of the \TopTen dataset (the token-based vocabulary stores \NTK unique tokens). Finally, we encode the method body by representing each token with its index in the token-based vocabulary.

\Part{Training Naive Models}.
We train 2-layer bi-directional GRUs \cite{cho2014gru} with PyTorch \footnote{\url{https://pytorch.org/docs/stable/generated/torch.nn.GRU.html}} on character-based representation (\BGCH) and token-based representation (\BGTK) for predicting the method name.
The classifier on \BGCH and \BGTK are referred to as \GRUCH and \GRUTK, respectively.

\subsection{Evaluation Metrics}
We use the following metrics as commonly used in the literature ~\cite{allamanis2017learning,alon2019code2vec} to evaluate the performance of handcrafted features. Suppose, $tp$ denotes the number of true positives, $tn$ denotes the number of true negatives, $fp$ denotes the number of false positives, and $fn$ denotes the number of false negatives in the results of the classification of a method on the test data.

\textit{Accuracy} indicates how many predicted examples are correct. It is the ratio of the correctly predicted examples to the total examples of the class.
\begin{align*} 
    Accuracy &= \frac{tp\,+\,tn}{tp\,+\,tn\,+\,fp\,+\,fn}
\end{align*}

\textit{Precision} indicates how many predicted examples are true positives. It is the ratio of the correctly predicted positive examples to the total predicted positive examples.
\begin{align*} 
    Precision &= \frac{tp}{tp\,+\,fp}
\end{align*}

\textit{Recall} indicates how many true positives examples are correctly predicted. It is the ratio of the correctly predicted positive examples to the total examples of the class.
\begin{align*} 
    Recall &= \frac{tp}{tp\,+\,fn}
\end{align*}

\textit{F1-Score} is the harmonic mean of precision (P) and recall (R).
\begin{align*} 
    F_{1}\text{--}Score &= \frac{2}{P^{-1} + R^{-1}} = 2\,.\,\frac{P\,.\,R}{P\,+\,R} 
\end{align*}

\section{Results}
\label{sec:results}

In this section, we will describe the experimental results including choice of handcrafted features, comparison of classifiers, and visualization.
Each classifier is trained on the corresponding training set, tuned on the validation set, and later evaluated on a separate test set.
In this section, the classifiers on \BGCH, \BGTK, \HCBCX, and \CtV feature vectors are referred to as \GRUCH, \GRUTK, \SVMHC, and \SVMNN, respectively.

	
	


\subsection{Choice of Handcrafted Features}
\label{subsec:result_hc}

\begin{table} 
    \centering
    \def\arraystretch{1.1}
    \caption{Result of handcrafted features on \TopTen dataset.}
    \label{table:result_hc}
    \resizebox{\columnwidth}{!}{%
    \begin{tabular}{|c|c|c|c|c|}
        \hline
        \textbf{Method} & \textbf{Feature Vectors} & \textbf{Precision} & \textbf{Recall} & \textbf{$F_1$-Score} \\ \hline
        \hline
        
        \multirow{4}{*}{equals}
        & \HCB   & 98.54 & 98.88 & 98.71 \\ \cline{2-5}
        & \HCN   & 98.20 & 97.98 & 98.09 \\ \cline{2-5}
        & \HCBCX & 99.21 & 98.54 & \textbf{98.87} \\ \cline{2-5}
        & \HCNCX & 98.99 & 98.76 & \textbf{98.87} \\ \hline
        \hline
        
        \multirow{4}{*}{main}
        & \HCB   & 94.62 & 96.85 & \underline{95.72} \\ \cline{2-5}
        & \HCN   & 91.70 & 94.59 & 93.12 \\ \cline{2-5}
        & \HCBCX & 94.72 & 97.15 & \textbf{95.92} \\ \cline{2-5}
        & \HCNCX & 91.04 & 94.98 & 92.97 \\ \hline
        \hline
        
        \multirow{4}{*}{setUp}
        & \HCB   & 87.70 & 86.10 & 86.89 \\ \cline{2-5}
        & \HCN   & 78.90 & 90.87 & 84.46 \\ \cline{2-5}
        & \HCBCX & 90.26 & 93.68 & \textbf{91.94} \\ \cline{2-5}
        & \HCNCX & 87.53 & 92.70 & \underline{90.04} \\ \hline
        \hline
        
        \multirow{4}{*}{onCreate}
        & \HCB   & 100.00 & 92.99 & \underline{96.37} \\ \cline{2-5}
        & \HCN   & 100.00 & 92.86 & 96.30 \\ \cline{2-5}
        & \HCBCX & 99.86  & 93.13 & \textbf{96.38} \\ \cline{2-5}
        & \HCNCX & 100.00 & 92.45 & 96.08 \\ \hline
        \hline
        
        \multirow{4}{*}{toString}
        & \HCB   & 93.41 & 97.65 & \textbf{95.48} \\ \cline{2-5}
        & \HCN   & 93.56 & 95.46 & 94.50 \\ \cline{2-5}
        & \HCBCX & 95.57 & 94.52 & \underline{95.04} \\ \cline{2-5}
        & \HCNCX & 94.81 & 94.37 & 94.59 \\ \hline
        \hline
        
        \multirow{4}{*}{run}
        & \HCB   & 62.03 & 61.87 & 61.95 \\ \cline{2-5}
        & \HCN   & 60.51 & 75.74 & 67.27 \\ \cline{2-5}
        & \HCBCX & 69.24 & 66.75 & \underline{67.97} \\ \cline{2-5}
        & \HCNCX & 69.55 & 70.09 & \textbf{69.82} \\ \hline
        \hline
        
        \multirow{4}{*}{hashCode}
        & \HCB   & 97.06 & 94.29 & 95.65 \\ \cline{2-5}
        & \HCN   & 96.85 & 95.84 & 96.34 \\ \cline{2-5}
        & \HCBCX & 98.95 & 97.92 & \textbf{98.43} \\ \cline{2-5}
        & \HCNCX & 98.19 & 98.44 & \underline{98.31} \\ \hline
        \hline
        
        \multirow{4}{*}{init}
        & \HCB   & 74.73 & 94.25 & \underline{83.36} \\ \cline{2-5}
        & \HCN   & 73.55 & 92.17 & 81.81 \\ \cline{2-5}
        & \HCBCX & 77.72 & 90.58 & \textbf{83.66} \\ \cline{2-5}
        & \HCNCX & 75.43 & 91.69 & 82.77 \\ \hline
        \hline
        
        \multirow{4}{*}{execute}
        & \HCB   & 76.25 & 86.89 & 81.22 \\ \cline{2-5}
        & \HCN   & 63.60 & 94.59 & 76.06 \\ \cline{2-5}
        & \HCBCX & 80.67 & 82.05 & \underline{81.35} \\ \cline{2-5}
        & \HCNCX & 76.36 & 92.02 & \textbf{83.46} \\ \hline
        \hline
        
        \multirow{4}{*}{get}
        & \HCB   & 86.76 & 95.82 & \underline{91.07} \\ \cline{2-5}
        & \HCN   & 84.96 & 91.04 & 87.89 \\ \cline{2-5}
        & \HCBCX & 89.89 & 95.52 & \textbf{92.62} \\ \cline{2-5}
        & \HCNCX & 88.54 & 92.24 & 90.35 \\ \hline

    \end{tabular}%
    }
\end{table}

Table~\ref{table:result_hc} shows the detailed result of handcrafted features on the \TopTen dataset where the \textbf{bold} values represent the best results and the \underline{underlined} values represent the second-best result. In this table, ``HC'' stands for handcrafted features. ``\HCB'' and ``\HCN'' denote the handcrafted features as binary vectors and numeric vectors, respectively. Similarly, ``\CX'' is to indicate the additional complexity features as numeric vectors.

\subsubsection{Binary vectors vs. Numeric vectors}
\label{subsec:result_vecs}
\begin{figure*}
\noindent \begin{minipage}{.44\textwidth}
\includegraphics[width=0.99\linewidth]{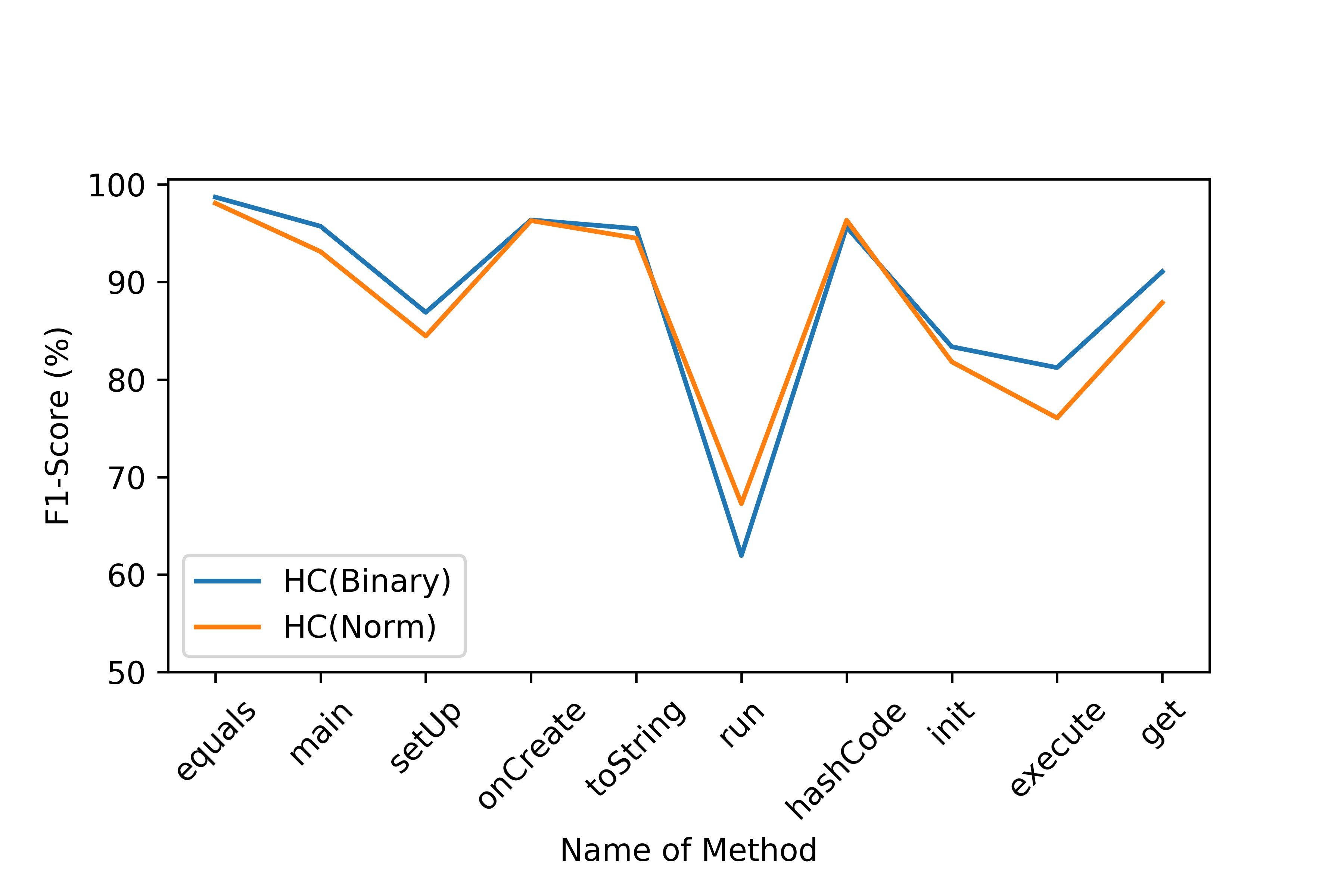}
\caption*{(a) Method-only features.}
\end{minipage}%
\begin{minipage}{.44\textwidth}
\includegraphics[width=0.99\linewidth]{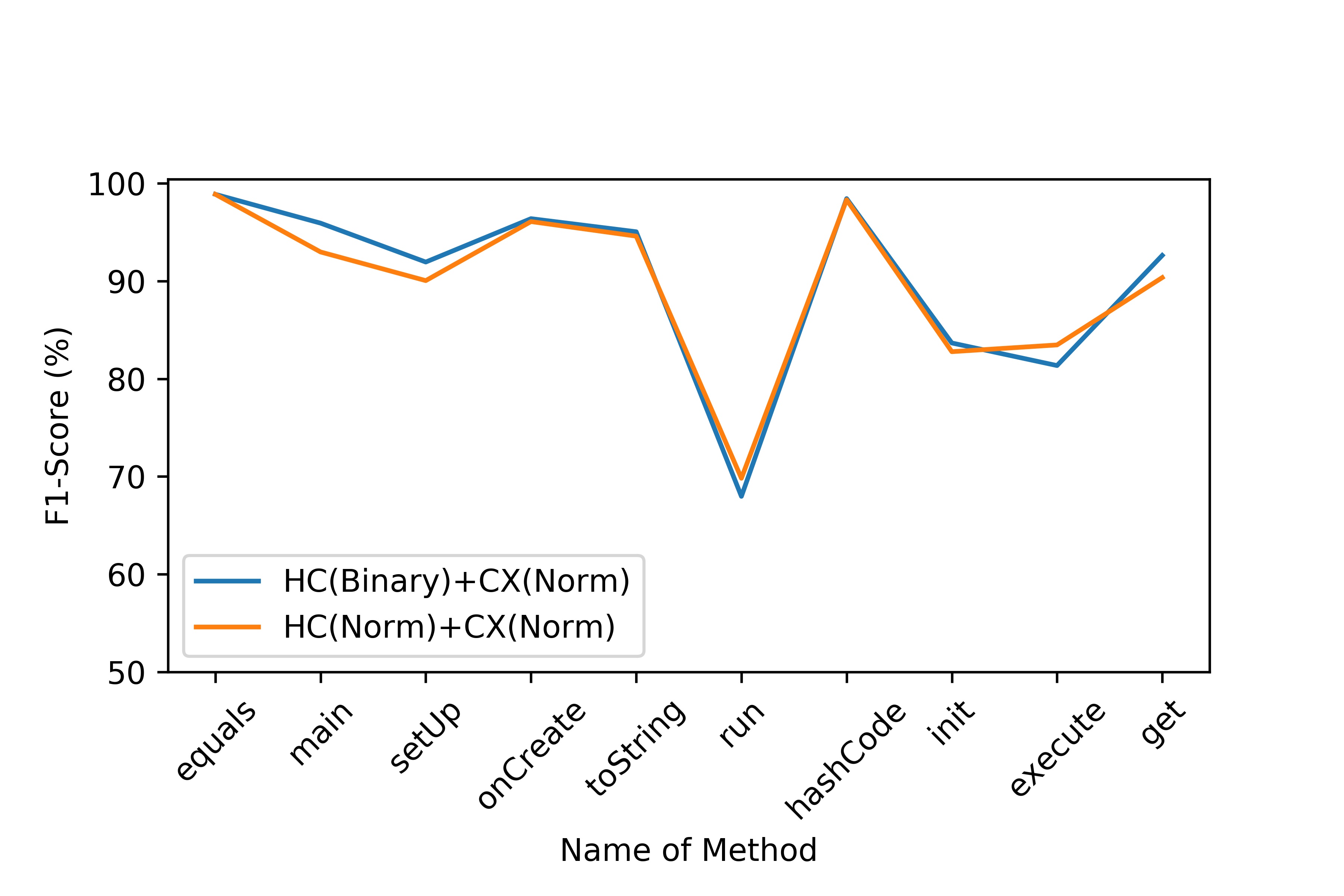}
\caption*{(b) Method+Complexity features.}
\end{minipage}
\caption{Binary vectors vs. Numeric vectors.}
\label{fig:binary_vs_numeric}
\end{figure*}

\begin{table}
    \begin{center}
        \def\arraystretch{1.1}
        \caption{Type of feature vectors.}
        \label{table:feature_types}
        \resizebox{\columnwidth}{!}{%
        \begin{tabular}{|c|p{5.8cm}|}
            \hline
            \textbf{Feature Vectors} & \textbf{Definition} \\
            \hline \hline 
                \BGCH & {A sequence of ASCII characters represented by its index in a character-based vocabulary.} \\ \hline
                \BGTK & {A sequence of Java tokens represented by its index in a token-based vocabulary.} \\ \hline
                \HCB & {The \NMF handcrafted features of methods as binary vectors.} \\ \hline
                \HCN & {The \NMF handcrafted features of methods as numeric vectors.} \\ \hline
                \scalebox{0.9}{\HCBCX} & {\HCB with the additional \NCF complexity features as numeric vectors.} \\ \hline
                \scalebox{0.9}{\HCNCX} & {\HCN with the additional \NCF complexity features as numeric vectors.} \\ \hline
                \CtV & {The code vectors of 384 dimensions from code2vec model \cite{alon2019code2vec}.} \\ \hline
        \end{tabular}%
        }
    \end{center}
\end{table}

\begin{table} 
    \centering
    \def\arraystretch{1.1}
    \caption{Average results on the \TopTen dataset.}
    \label{table:result_avg}
    \resizebox{0.98\columnwidth}{!}{%
    \begin{tabular}{|c|c|c|c|c|}
        \hline
        \textbf{Feature Vectors} & \textbf{Accuracy} & \textbf{Precision} & \textbf{Recall} & \textbf{$F_1$-Score} \\ \hline
        \hline

        \BGCH & 38.65 & 26.02 & 38.65 & 30.57 \\  \hline
        \BGTK & 70.58 & 60.38 & 70.59 & 63.37 \\  \hline
        \HCB & 88.32 & 87.11 & 90.56 & 88.64 \\  \hline
        \HCN & 86.27 & 84.18 & 92.11 & 87.58 \\  \hline
        \HCBCX & 90.14 & 89.61 & 90.98 & \underline{90.22} \\  \hline
        \HCNCX & 89.36 & 88.04 & 91.77 & 89.73 \\  \hline
        \CtV & 93.73 & 95.54 & 91.38 & \textbf{93.24} \\ \hline

    \end{tabular}%
    }
\end{table}

Figure~\ref{fig:binary_vs_numeric} depicts how the choice of presence (binary vectors) or number of occurrences (numeric vectors) influences the quality of handcrafted features. 
We compare binary vectors and numeric vectors on (a) method-only features: \HCB vs. \HCN, and (b) method+complexity features: \HCBCX vs. \HCNCX.
In Figure \ref{fig:binary_vs_numeric}(a) and \ref{fig:binary_vs_numeric}(b), the blue line shows the $F_1$-Score when the features are considered as binary vectors and the orange line shows the $F_1$-Score when the features are considered as numeric vectors. 
According to Figure~\ref{fig:binary_vs_numeric}(a), in most cases, the \HCB are comparatively better than the \HCN except for the `run' and `hashCode' methods where the difference are 5.32\% and 0.69\%, respectively.
Similarly, in most cases, the \HCBCX are comparatively better than the \HCNCX in Figure~\ref{fig:binary_vs_numeric}(b) except for the `run' and `execute' methods where the difference are $1.85\%$ and $2.11\%$, respectively.
The average $F_1$-Score of Table~\ref{table:result_avg} also shows that the \HCB is almost 1\% better than the \HCN and the \HCBCX is almost 0.5\% better than the \HCNCX.
This can suggest that only the presence of features can be used to recognize a method, instead of counting the number of occurrences of features.

\observation{The presence of a feature can be used to recognize a method instead of counting the number of occurrences of that feature in programs. On average, the choice of binary vectors has increased the $F_1$-Score up to $1\%$ than the numeric vectors.}

\subsubsection{Impact of Additional Complexity Features.}
\label{subsec:result_com}

\begin{figure*}
\noindent \begin{minipage}{.44\textwidth}
\includegraphics[width=0.99\linewidth]{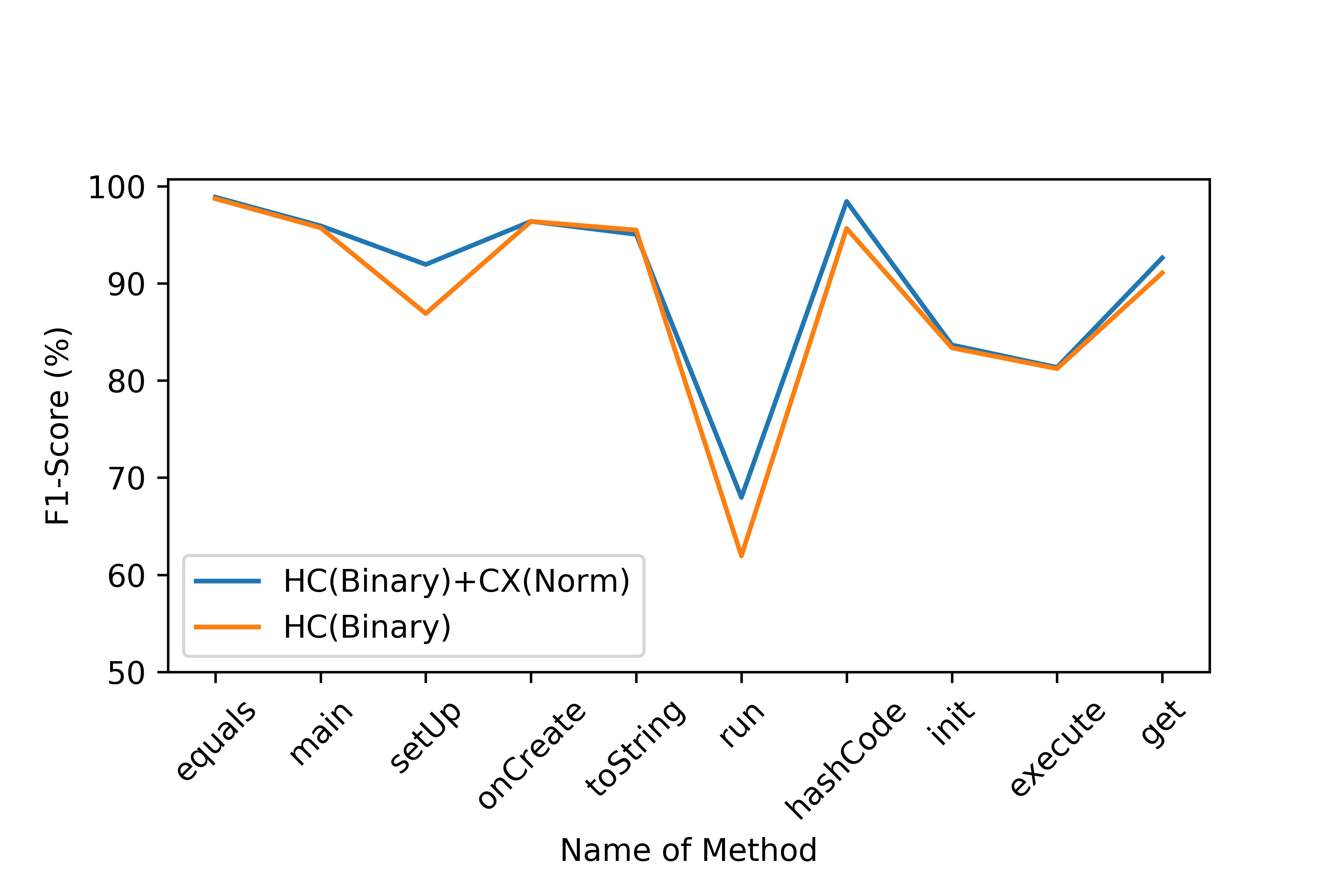}
\caption*{(a) Binary vectors.}
\end{minipage}%
\begin{minipage}{.44\textwidth}
\includegraphics[width=0.99\linewidth]{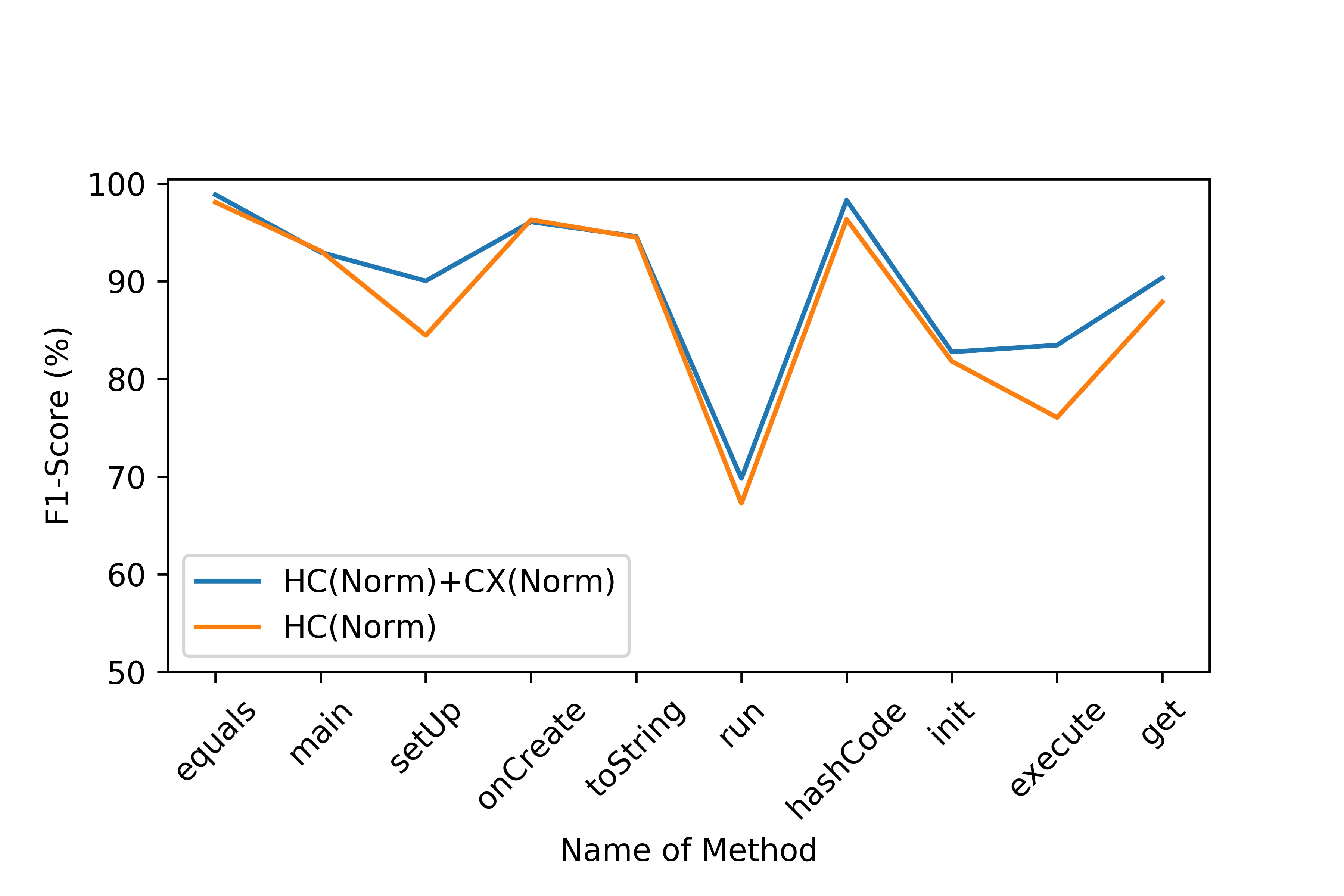}
\caption*{(b) Numeric vectors.}
\end{minipage}
\caption{Impact of additional complexity features.}
\label{fig:add_com_features}
\end{figure*}

Figure~\ref{fig:add_com_features} depicts the importance of complexity features on the quality of handcrafted features. 
We compare method-only features and  method+complexity features on (a) binary vectors: \HCB vs. \HCBCX, and (b) numeric vectors: \HCN vs. \HCNCX.
In Figure \ref{fig:add_com_features}(a) and \ref{fig:add_com_features}(b), the blue line and orange line shows the $F_1$-Score of method-only features and method+complexity features, respectively. 
According to Figure~\ref{fig:add_com_features}(a), in most cases, the \HCBCX are comparatively better than the \HCB except for the `toString' method where the difference is $0.44\%$.
Similarly, in most cases, the \HCNCX are comparatively better than the \HCN in Figure~\ref{fig:add_com_features}(b) except for the `main' and `onCreate' methods where the difference are 0.15\% and 0.22\%, respectively.
The average $F_1$-Score of Table~\ref{table:result_avg} also shows that the \HCBCX is almost $1.6\%$ better than the \HCB and the \HCNCX is almost $2.2\%$ better than the \HCN.
This can suggest that the code complexity features can be useful to better recognize a method, especially for some methods (i.e. `setUp', `run', `hashCode', and `execute') where the improvements for additional code complexity features are almost $3\sim7\%$.

\observation{The code complexity features can be useful to better recognize a method along with the method-only features. On average, the additional code complexity features have increased the $F_1$-Score up to $2.2\%$ than the method-only features.}

\begin{table}
    \centering
    \def\arraystretch{1.1}
    \caption{Result of different feature vectors on \TopTen dataset.}
    \label{table:result_clf}
    \resizebox{\columnwidth}{!}{%
    \begin{tabular}{|c|c|c|c|c|}
        \hline
        \textbf{Method} & \textbf{Feature Vectors} & \textbf{Precision} & \textbf{Recall} & \textbf{$F_1$-Score} \\ \hline
        \hline
        
        \multirow{4}{*}{equals}
        & \BGCH  & 50.97 & 74.02 & 60.37 \\ \cline{2-5}
        & \BGTK  & 99.20 & 97.53 & 98.36 \\ \cline{2-5}
        & \HCBCX & 99.21 & 98.54 & \underline{98.87} \\ \cline{2-5}
        & \CtV   & 99.55 & 99.10 & \textbf{99.32} \\ \hline
        \hline
        
        \multirow{4}{*}{main}
        & \BGCH  &  0.00 &  0.00 & 0.00 \\ \cline{2-5}
        & \BGTK  & 84.38 & 65.94 & 74.03 \\ \cline{2-5}
        & \HCBCX & 94.72 & 97.15 & \underline{95.92} \\ \cline{2-5}
        & \CtV   & 98.72 & 98.52 & \textbf{98.62} \\ \hline
        \hline
        
        \multirow{4}{*}{setUp}
        & \BGCH  & 26.12 & 59.83 & 36.36 \\ \cline{2-5}
        & \BGTK  & 42.93 & 89.19 & 57.96 \\ \cline{2-5}
        & \HCBCX & 90.26 & 93.68 & \underline{91.94} \\ \cline{2-5}
        & \CtV   & 99.26 & 94.10 & \textbf{96.61} \\ \hline
        \hline
        
        \multirow{4}{*}{onCreate}
        & \BGCH  & 59.89 & 87.74 & 71.19 \\ \cline{2-5}
        & \BGTK  & 94.70 & 91.51 & 93.08 \\ \cline{2-5}
        & \HCBCX & 99.86 & 93.13 & \underline{96.38} \\ \cline{2-5}
        & \CtV  & 100.00 & 99.06 & \textbf{99.53} \\ \hline
        \hline
        
        \multirow{4}{*}{toString}
        & \BGCH  & 51.64 & 74.02 & 60.84 \\ \cline{2-5}
        & \BGTK  & 85.14 & 88.73 & 86.90 \\ \cline{2-5}
        & \HCBCX & 95.57 & 94.52 & \underline{95.04} \\ \cline{2-5}
        & \CtV   & 97.37 & 98.44 & \textbf{97.90} \\ \hline
        \hline
        
        \multirow{4}{*}{run}
        & \BGCH  & 25.36 & 27.47 & 26.37 \\ \cline{2-5}
        & \BGTK  & 37.96 & 51.99 & 43.88 \\ \cline{2-5}
        & \HCBCX & 69.24 & 66.75 & \underline{67.97} \\ \cline{2-5}
        & \CtV   & 86.30 & 62.26 & \textbf{72.33} \\ \hline
        \hline
        
        \multirow{4}{*}{hashCode}
        & \BGCH  & 30.18 & 52.99 & 38.45 \\ \cline{2-5}
        & \BGTK  & 74.70 & 97.40 & 84.55 \\ \cline{2-5}
        & \HCBCX & 98.95 & 97.92 & \underline{98.43} \\ \cline{2-5}
        & \CtV   & 99.74 & 99.74 & \textbf{99.74} \\ \hline
        \hline
        
        \multirow{4}{*}{init}
        & \BGCH  &  0.00 &  0.00 &  0.00 \\ \cline{2-5}
        & \BGTK  &  0.00 &  0.00 &  0.00 \\ \cline{2-5}
        & \HCBCX & 77.72 & 90.58 & \underline{83.66} \\ \cline{2-5}
        & \CtV   & 88.74 & 87.54 & \textbf{88.14} \\ \hline
        \hline
        
        \multirow{4}{*}{execute}
        & \BGCH  &  2.44 &  0.28 &  0.51 \\ \cline{2-5}
        & \BGTK  & 41.04 & 31.34 & 35.54 \\ \cline{2-5}
        & \HCBCX & 80.67 & 82.05 & \underline{81.35} \\ \cline{2-5}
        & \CtV   & 93.44 & 85.19 & \textbf{89.12} \\ \hline
        \hline
        
        \multirow{4}{*}{get}
        & \BGCH  & 13.55 & 10.15 & 11.60 \\ \cline{2-5}
        & \BGTK  & 43.77 & 92.24 & 59.37 \\ \cline{2-5}
        & \HCBCX & 89.89 & 95.52 & \textbf{92.62} \\ \cline{2-5}
        & \CtV   & 92.33 & 89.85 & 91.07 \\ \hline

    \end{tabular}%
    }
\end{table}


\subsection{Comparison of Classifiers}
\label{subsec:result_clf}

\begin{figure*}
\noindent \begin{minipage}{.5\textwidth}
\includegraphics[width=0.99\linewidth]{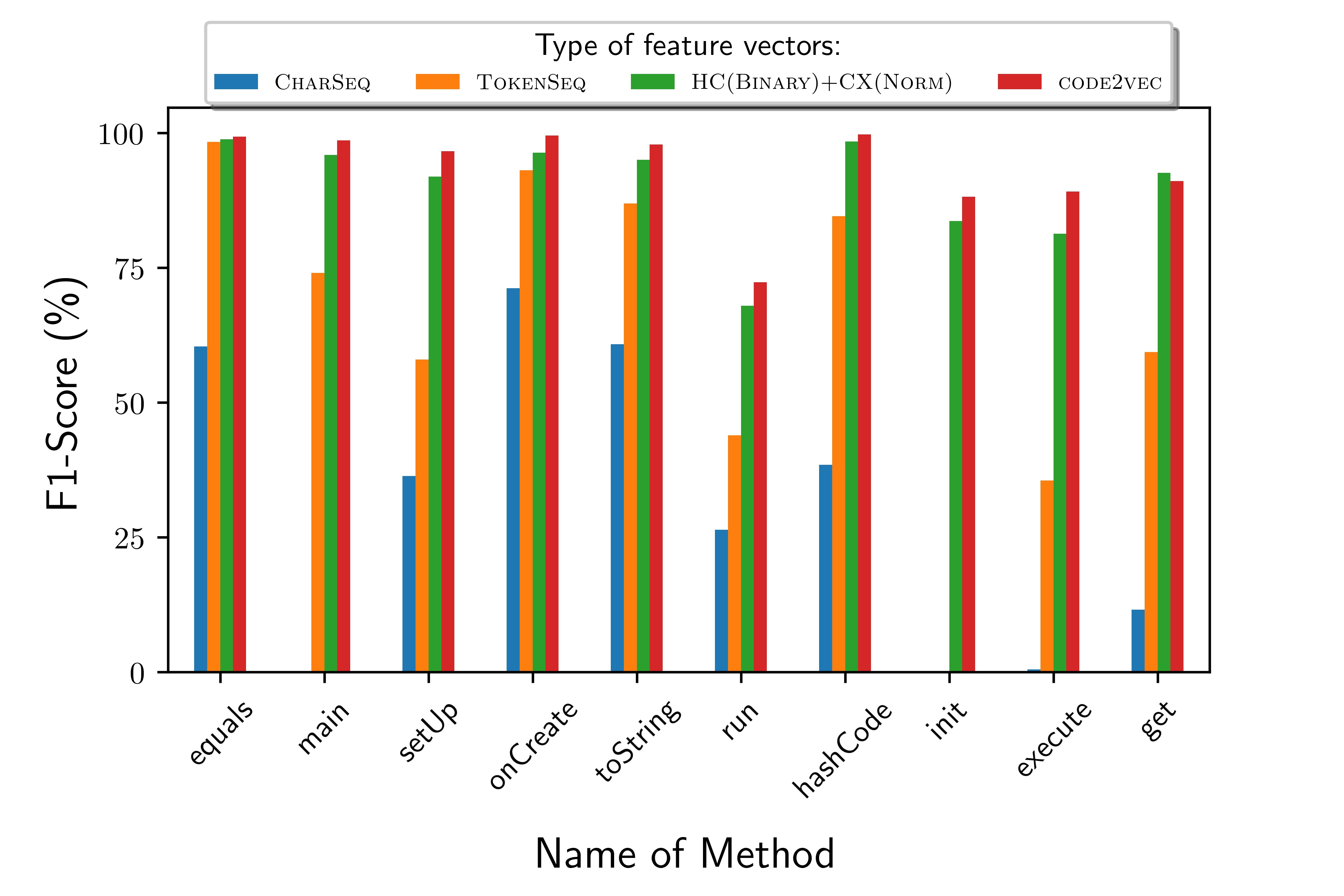}
\caption*{(a) Barplots over method names.}
\end{minipage}%
\begin{minipage}{.5\textwidth}
\includegraphics[width=0.99\linewidth]{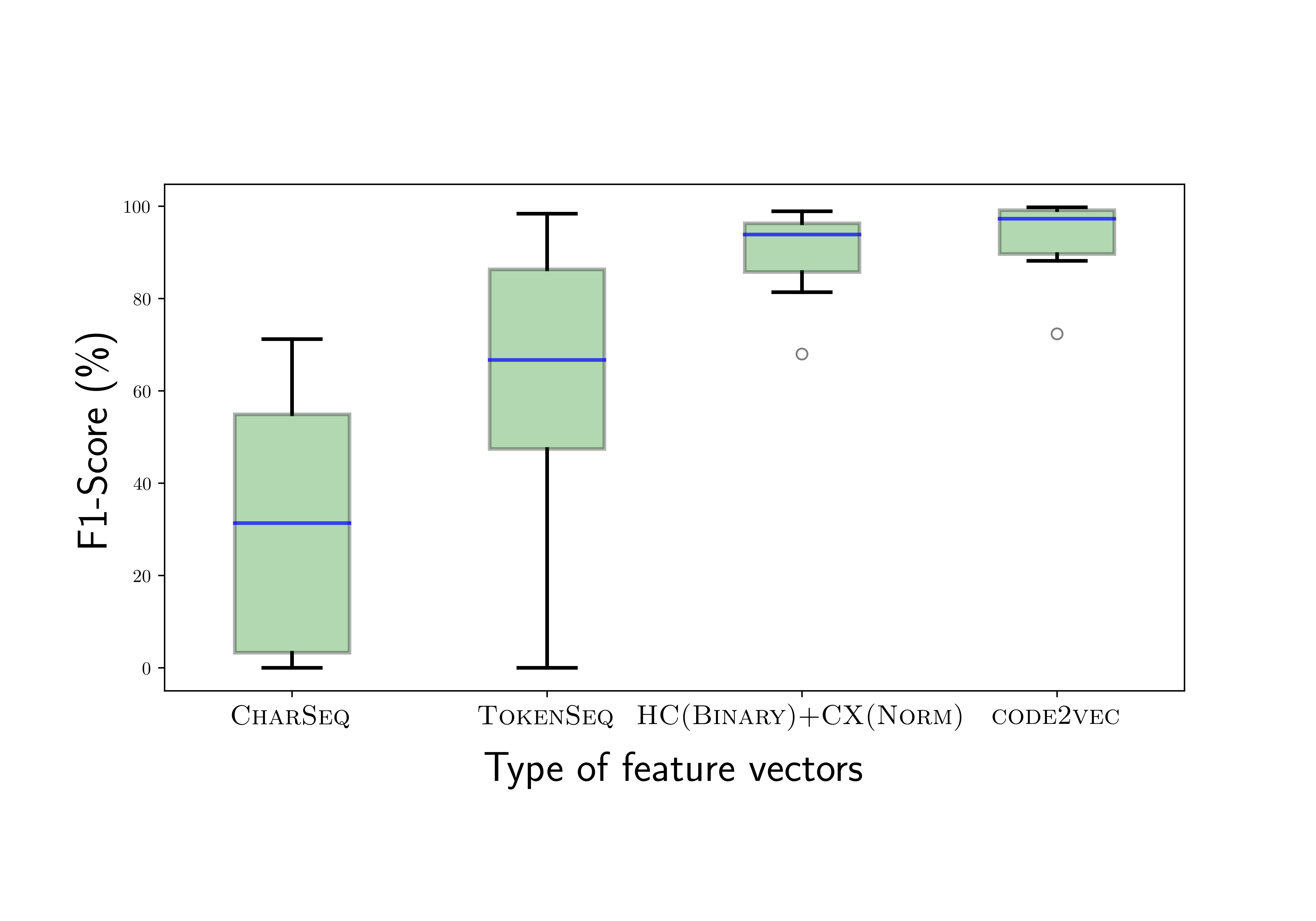}
\caption*{(b) Boxplots over feature vectors.}
\end{minipage}
\caption{Comparison of classifiers on \TopTen dataset.}
\label{fig:hc_c2v}
\end{figure*}

Table~\ref{table:result_clf} shows the detailed result of different feature vectors on the \TopTen dataset where the \textbf{bold} values represent the best results and the \underline{underlined} values represent the second-best result.
We also draw the commonly used explanatory data plots (barplots over method names in Figure~\ref{fig:hc_c2v}a and boxplots over feature vectors in Figure~\ref{fig:hc_c2v}b) to visually show the distribution of results on the \TopTen dataset. 
As shown in the previous section (Section~\ref{subsec:result_hc}), in most cases, the binary vectors perform relatively better than the numeric vectors, and the code complexity features also improve the performance for handcrafted features. 
Therefore, in this section, we mainly compare the result of \HCBCX from handcrafted features. 

\subsubsection{\SVMHC vs. Sequence-based Baselines.}
In this section, we compare our handcrafted features against the following two sequence-based baselines: (a) a sequence of ASCII characters (\BGCH), and (b) a sequence of Java tokens (\BGTK).
According to Table~\ref{table:result_clf} and Figure~\ref{fig:hc_c2v}a, for all methods, our \SVMHC outperforms both \GRUCH and \GRUTK by a large margin for predicting method name. 
Even in some cases, \GRUCH (i.e. main, init, and execute) and \GRUTK (i.e. init) fail to predict the method name. 
The boxplots in Figure~\ref{fig:hc_c2v}b indicates that the variance of $F_1$-Scores among methods are also very significant for \GRUCH and \GRUTK. 
The average $F_1$-Score of Table~\ref{table:result_avg} also shows that the \SVMHC is $59.65\%$ and $26.85\%$ better than the \GRUCH and the \GRUTK, respectively.

\observation{The handcrafted features significantly outperform the sequence of characters (by $59.65\%$) and the sequence of tokens (by 26.85\%) for predicting method name.}

\subsubsection{\SVMHC vs. \SVMNN.}
In this section, we compare our handcrafted features with the path-based embedding of code2vec \cite{alon2019code2vec}.
According to Table~\ref{table:result_clf} and Figure~\ref{fig:hc_c2v}, the \SVMNN performs better than the \SVMHC but the difference is not always significant. 
When the $F_1$-Score of \SVMNN is near perfect (i.e., equals, onCreate, and hashCode), the $F_1$-Score of \SVMHC is also higher and very close to the \SVMNN. 
Similarly, for some other methods (i.e., run, init, and execute), they both perform relatively worst. 
However, there are some cases where the difference between \SVMNN and \SVMHC is  significant, for example, \SVMNN shows almost 8\% improvement over \SVMHC to classify the `execute' method.
On the other hand, \SVMHC is 1.5+\% better than \SVMNN to classify the `get' method.
The average $F_1$-Score also shows that the \SVMNN obtains around 3\% improvements over the \SVMHC.
This can suggest that the handcrafted features with a very smaller feature set can achieve highly comparable results to the higher dimensional embeddings of deep neural model such as \CtV.

\observation{The handcrafted features with a very smaller feature set can achieve highly comparable results to the higher dimensional embeddings of deep neural model such as \CtV.}

\subsection{Information Gains and Importance of Dimensions}
\label{subsec:result_ig}
\begin{figure}
    \centering
    \includegraphics[width=\columnwidth]{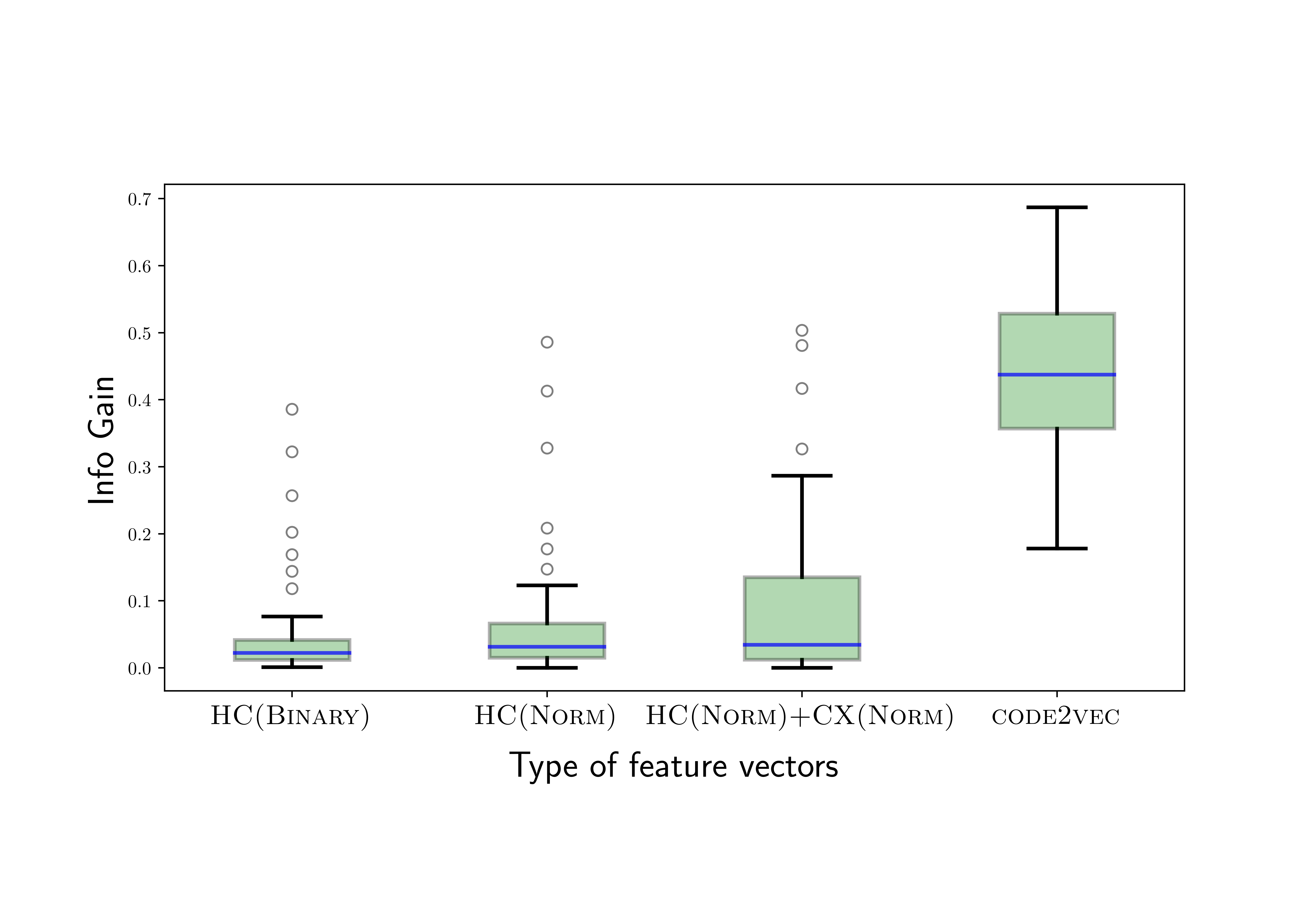}
    \caption{The distribution of information gain for `equals' method.}
    \label{fig:equals_ig}
\end{figure}

Figure~\ref{fig:equals_ig} depicts the distribution of information gains of each dimension, i.e., feature, in the `equals' dataset. 
It suggests that the information gains of features in code2vec embeddings is on average higher than the information gains of features in handcrafted features. However, the distribution of information gains in code2vec embeddings is symmetric while is highly skewed in handcrafted features.

We used the information gains and created new SVM models for methods such as `main' and `setUp' by using features with top $25\%$ of information gains.
The $F_1$-score of SVM models for binary handcrafted features (\HCB) with features of top $25\%$ information gains were $93.5\%$ and $80.11\%$, for `main' and `setUp', respectively, while these value for top $25\%$ code2vec dimensions were $98.62\%$ and $96.28\%$, respectively.
It shows that the handcrafted features suffered a higher loss of performance than their code2vec embeddings counterparts.
It may suggest that a large portion of code2vec embeddings might be unnecessary for the acceptable classification, hence, the size of embedding can be reduced. 

\observation{Compare to the handcrafted features, the information gains are more evenly distributed in the code2vec embeddings. Moreover, the code2vec embeddings are more resilient to the removal of dimensions with low information gains than the handcrafted features.}

\subsection{Visualization of Feature Vectors}
\label{subsec:result_tsne}
\begin{figure*} 
\centering
\captionsetup{justification=centering}
\noindent \begin{minipage}{.33\textwidth}
\includegraphics[width=0.98\columnwidth]{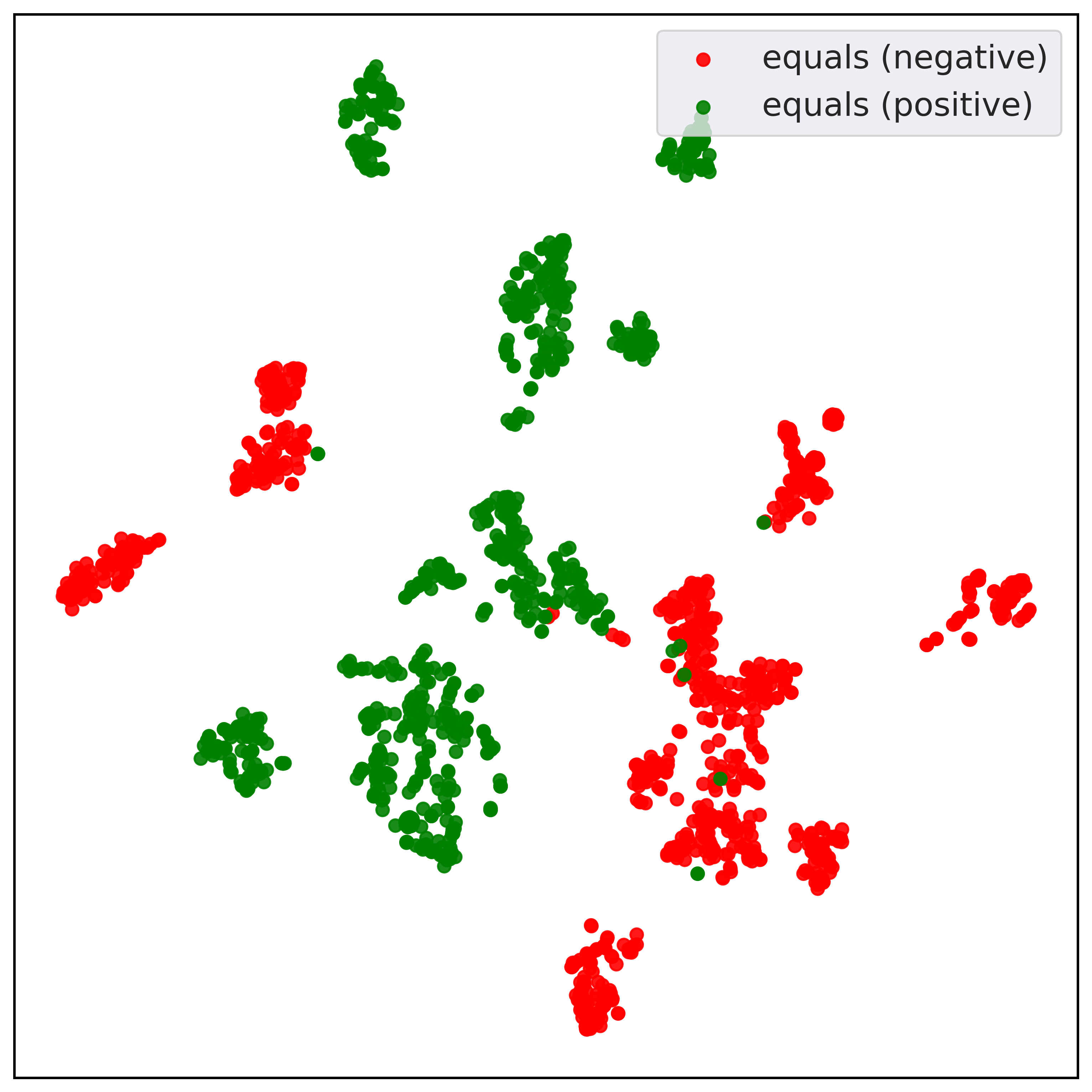}
\caption*{(a) \fsize{8}{\CtV [$F_1$ = 99.32\%]}}
\end{minipage}%
\begin{minipage}{.33\textwidth}
\includegraphics[width=0.98\columnwidth]{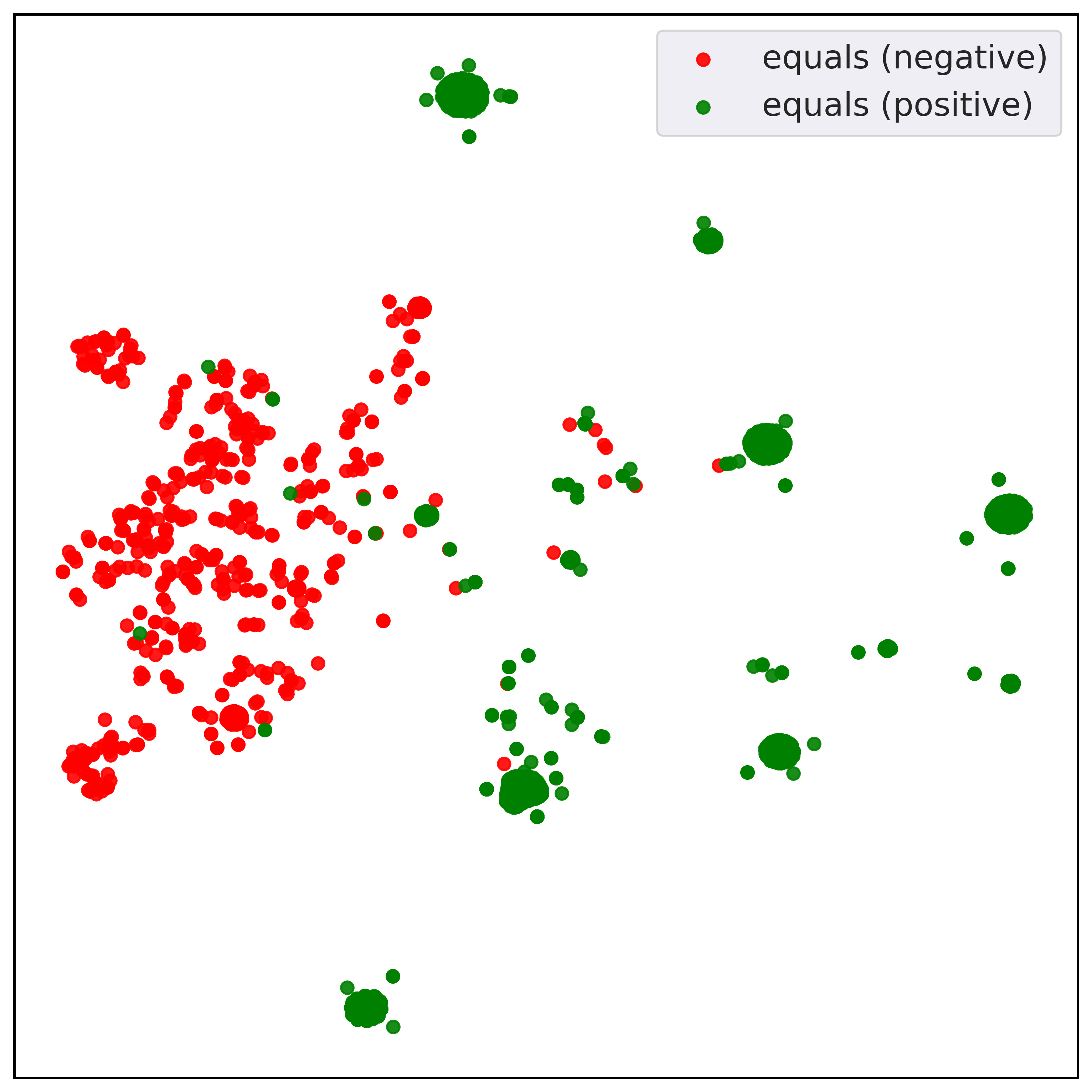}
\caption*{(b) \fsize{8}{\HCB [$F_1$ = 98.71\%]}}
\end{minipage}%
\begin{minipage}{.33\textwidth}
\includegraphics[width=0.98\columnwidth]{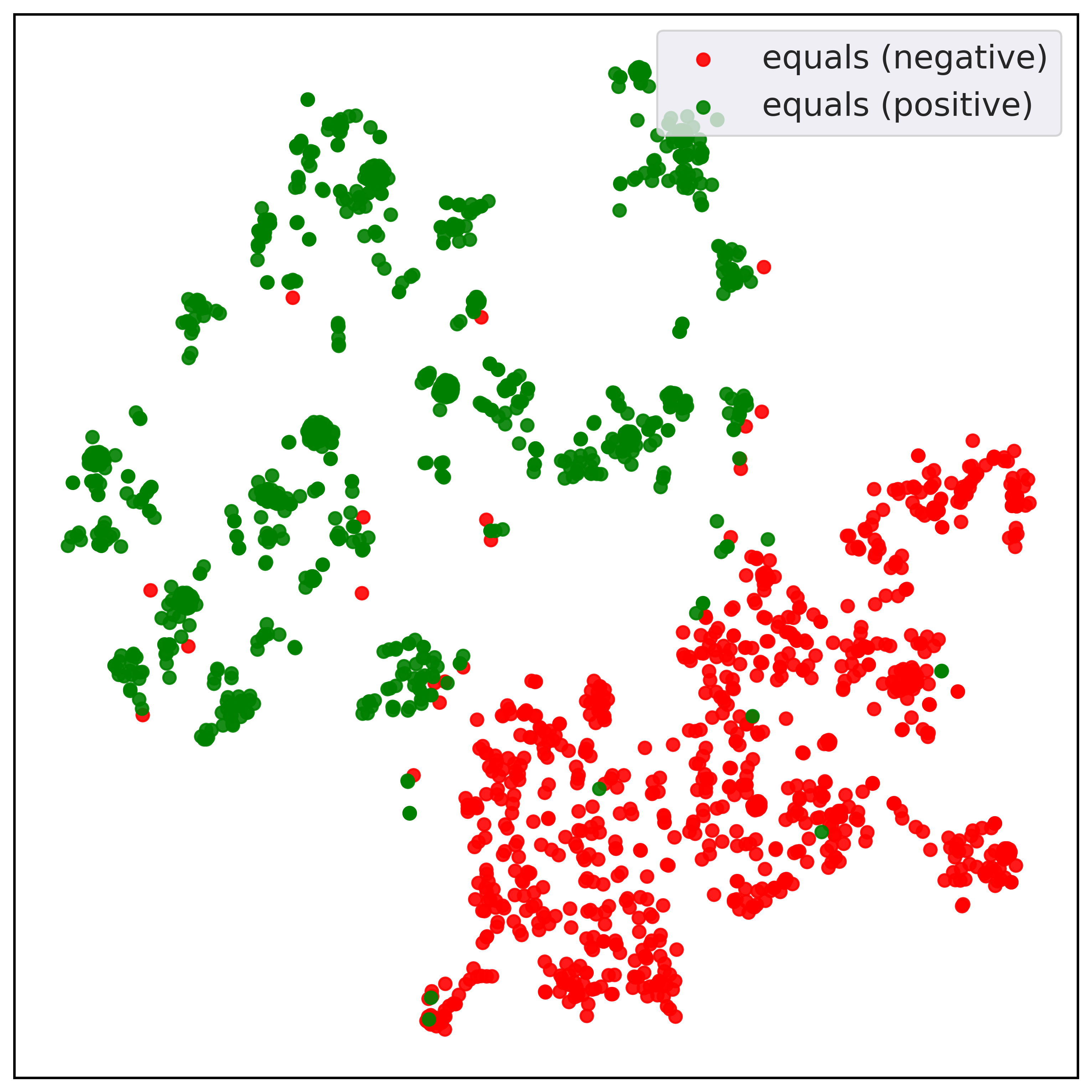}
\caption*{(c) \fsize{8}{\HCBCX [$F_1$ = 98.87\%]}}
\end{minipage}
\caption{The t-SNE plot of the best `equals' method.}
\label{fig:tsne_equals}
\end{figure*}

\begin{figure*} 
\centering
\captionsetup{justification=centering}
\noindent \begin{minipage}{.33\textwidth}
\includegraphics[width=0.98\columnwidth]{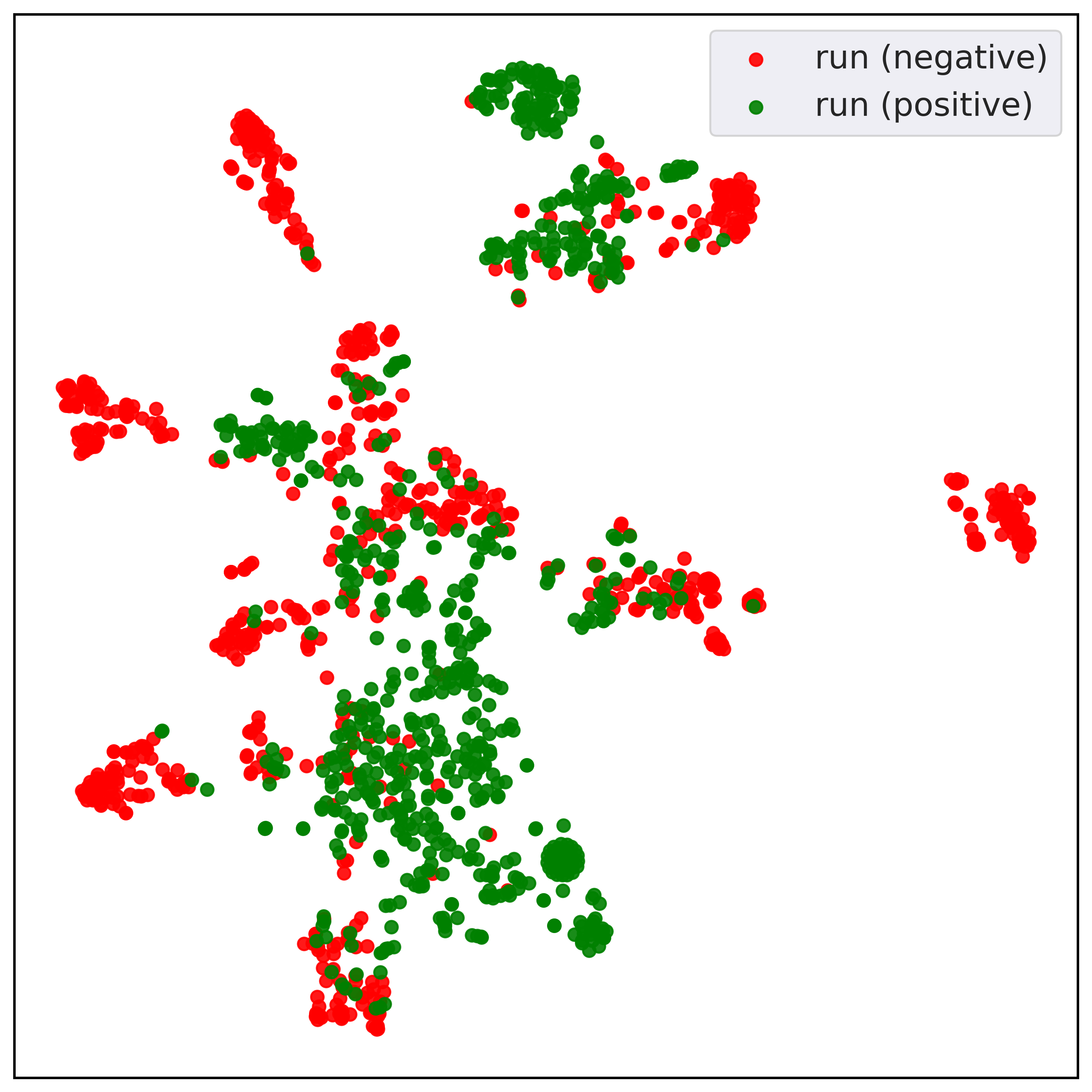}
\caption*{(a) \fsize{8}{\CtV [$F_1$ = 72.33\%]}}
\end{minipage}%
\begin{minipage}{.33\textwidth}
\includegraphics[width=0.98\columnwidth]{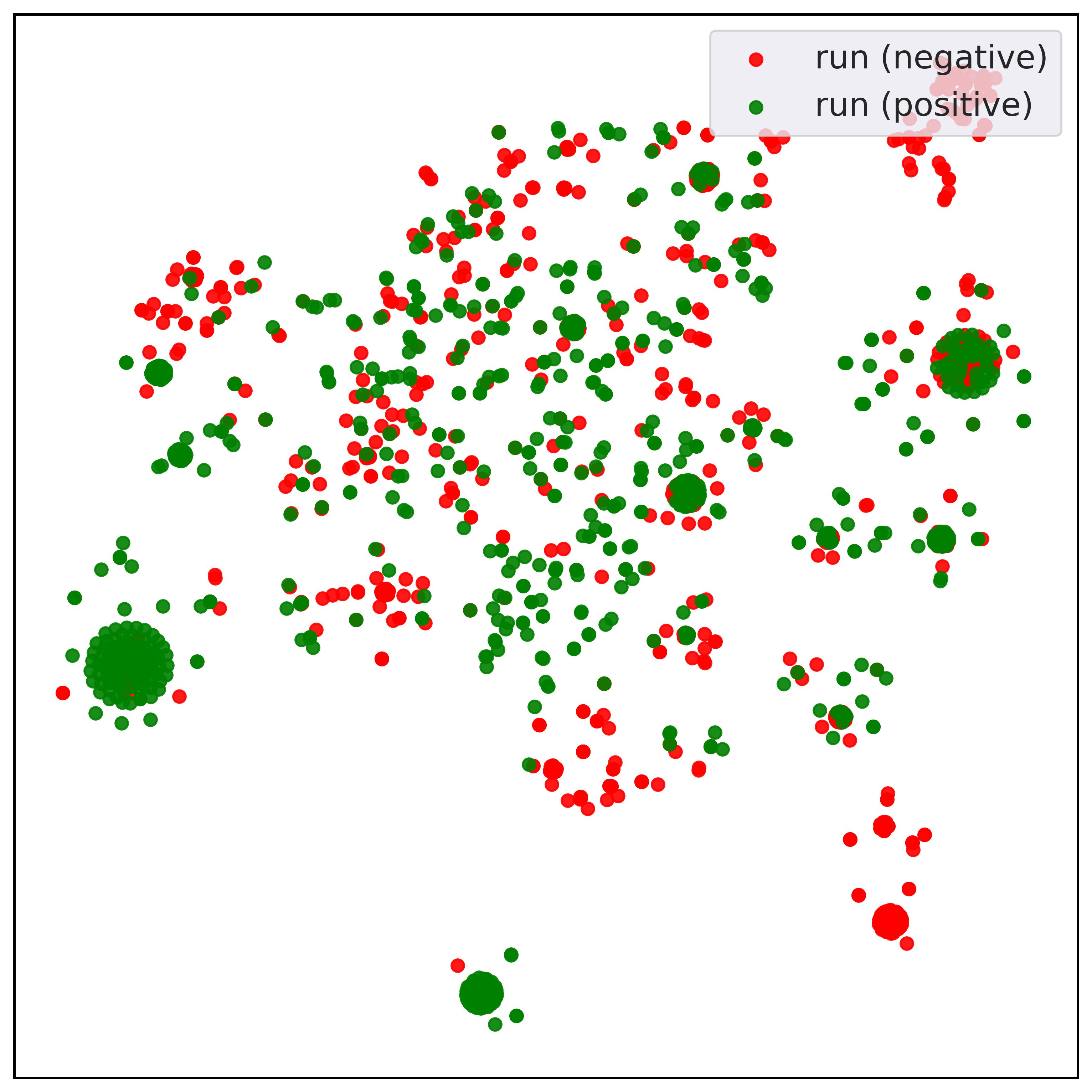}
\caption*{(b) \fsize{8}{\HCB [$F_1$ = 61.95\%]}}
\end{minipage}%
\begin{minipage}{.33\textwidth}
\includegraphics[width=0.98\columnwidth]{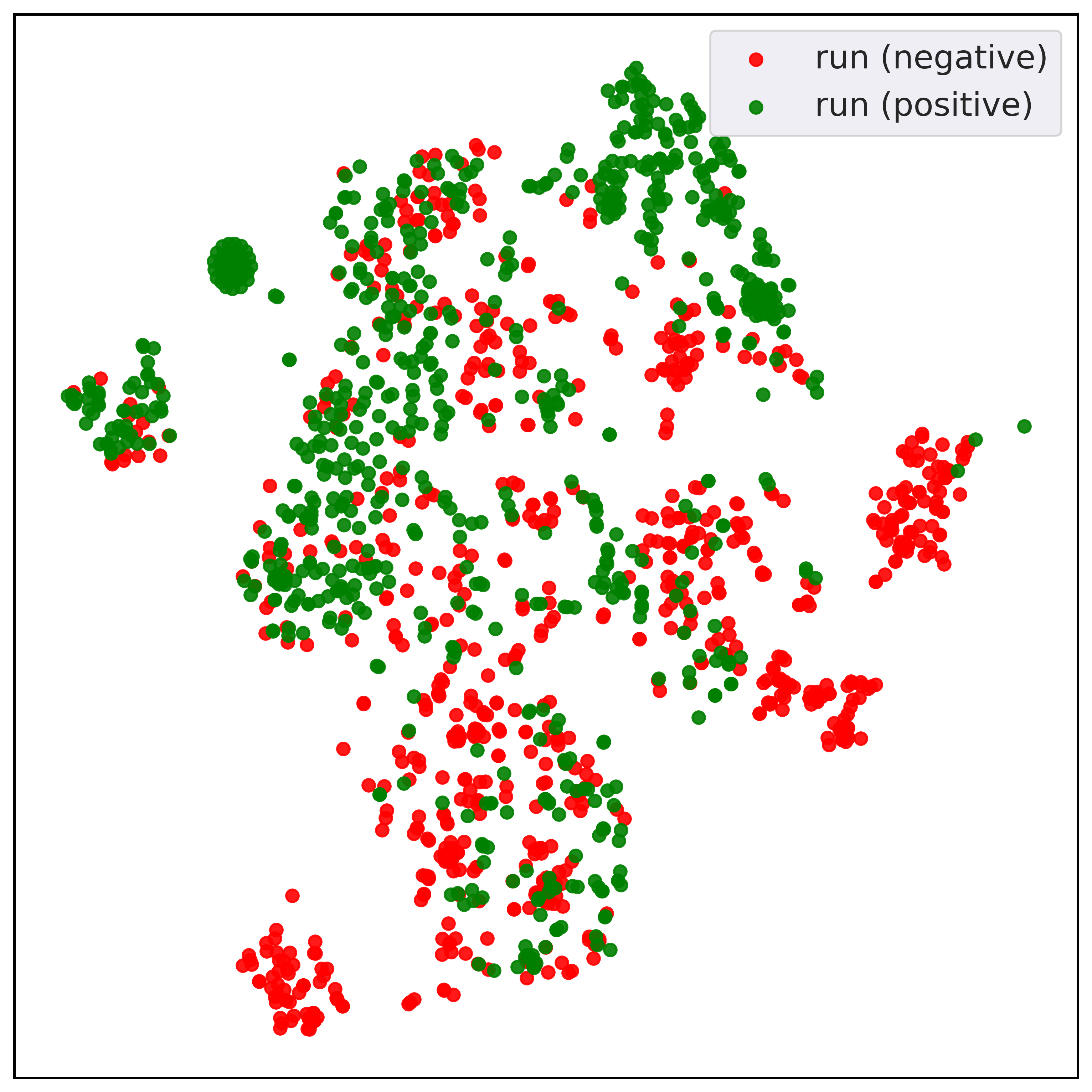}
\caption*{(c) \fsize{8}{\HCBCX [$F_1$ = 67.97\%]}}
\end{minipage}
\caption{The t-SNE plot of the worst `run' method.}
\label{fig:tsne_run}
\end{figure*}

To better understand how the features separate the positive and negative examples in the dataset we used t-SNE~\cite{maaten2008tsne} to project the feature vectors in code2vec embeddings and handcrafted features into two-dimensional space.
For illustration, we only visualize the method with best performing classifiers (i.e. `equals') in Figure \ref{fig:tsne_equals} and the method with worst performance classifiers (i.e. `run') in Figure \ref{fig:tsne_run}, for the \CtV, \HCB and \HCBCX, respectively.
Points from the same color (positive examples are in green color and negative examples are in red color) should tend to be grouped close to one another.

\Part{`equals' method}. Figure \ref{fig:tsne_equals} indicates that the data points are generally well grouped for the best method (`equals') where the positive points are quite distinct from the negative points.
The data points form a cluster of positive points in the middle of Figure \ref{fig:tsne_equals}a and are almost linearly separable in Figure \ref{fig:tsne_equals}b and \ref{fig:tsne_equals}c.
All show a good measure of separability as the $F_1$-Scores are nearly 100\%.

\Part{`run' method}. 
Figure \ref{fig:tsne_run} indicates that the data points are hardly separable.
The $F_1$-Score of Figure~\ref{fig:tsne_run}a is around 10\% higher than Figure~\ref{fig:tsne_run}b, thus the data points appear more scattered in Figure~\ref{fig:tsne_run}b than in Figure~\ref{fig:tsne_run}a. 
Similarly, the $F_1$-Score of Figure~\ref{fig:tsne_run}c is around 6\% higher than Figure~\ref{fig:tsne_run}b, thus the data points in Figure~\ref{fig:tsne_run}c are relatively less  scattered than in Figure~\ref{fig:tsne_run}b.

Although t-SNE plots are not objective ways to compare two embeddings, it may provide an intuition about the separability of methods based on the corresponding feature embeddings.
The figures \emph{might} suggest that the high-dimensional code2vec tends to produce a more complex hypothesis class than necessary, compared to the handcrafted features. 
Using too complex hypothesis class may increase the chances of overfitting in training the models.

\section{Related Work}
\label{sec:related}
Many studies have been done on the representation of source code \cite{BigCodeSurvey, chen2019literature} in machine learning models for predicting properties of programs such as identifier or variable names \cite{allamanis2014conventions, raychev2015crf, allamanis2017learning, alon2018path}, method names \cite{allamanis2015method, allamanis2016convolutional, alon2018path, allamanis2019structured, alon2019code2vec, alon2019code2seq, wang2020liger}, class names \cite{allamanis2015method}, types \cite{raychev2015crf, alon2018path, allamanis2018type}, and descriptions \cite{allamanis2019structured, alon2019code2seq}.
\citet{allamanis2014conventions} introduced a framework that processed token sequences and abstract syntax trees of code to suggest natural identifier names and formatting conventions on a Java corpus.
\citet{allamanis2015method} proposed a neural probabilistic language model with manually designed features from Java projects for suggesting method names and class names. 
\citet{raychev2015crf} converted the program into dependency representation that captured relationships between program elements and trained a CRF model for predicting the name of identifiers and predicting the type annotation of variables in JavaScript dataset.
\citet{allamanis2016convolutional} introduced a convolutional attention model for the code summarization task such as method name prediction with a sequence of subtokens from Java projects.
\citet{alon2018path} used the AST-based representation for learning properties of Java programs such as predicting variable names, predicting method names, and predicting full types. 
\citet{allamanis2017learning} constructed graphs from source code that leveraged data flow and control flow for predicting variable names and detecting variable misuses in C\# projects.
\citet{allamanis2018type} proposed a RNN-based model using sequence-to-sequence type annotations for type suggestion in TypeScript and plain JavaScript code.
\citet{allamanis2019structured} combined sequence encoders with graph neural networks that inferred relations among program elements for predicting name and description of the method in Java and C\# projects.
\citet{alon2019code2vec} used a bag of path-context from abstract syntax tree to learn the body of method for predicting the method name of Java projects. 
\citet{alon2019code2seq} later used an encoder-decoder architecture to encode the path-context as node-by-node to predict the method name of Java projects and the code caption of C\# projects.
\citet{liu2019inconsistent} used similar method bodies to spot and refactor inconsistent method names.
\citet{wang2020liger} embedded the symbolic and concrete execution traces of Java projects to learn program representations for method name prediction and semantics classification.

Apart from that, various deep neural embeddings and models have been also applied to different program analysis or software engineering tasks such as 
HAGGIS for mining idioms from source code \cite{allamanis2014haggis},
Gemini for binary code similarity detection \cite{xu2017gemini}, 
Code Vectors for code analogies, bug fining and repair/suggestion \cite{henkel2018code}, 
Dynamic Program Embeddings for classifying the types of errors in programs \cite{wang2018dynamic},
DYPRO for recognizing loop invariants \cite{wang2019dypro},
Import2Vec for learning embeddings of software libraries \cite{theeten2019import2vec},
NeurSA for catching static bugs in code \cite{wang2019NeurSA}, 
and HOPPITY to detect and fix bugs in programs \cite{dinella2020hoppity}.
Researchers have also studied the language model for code completion \cite{hindle2012ngram, raychev2014methodcall, raychev2016dt}, code suggestion \cite{allamanis2013gigatoken}, and code retrieval \cite{iyer2016lstm} task.

Moreover, \citet{BigCodeSurvey} survey the taxonomy of probabilistic models of source code and their applications, \citet{jiang2019recommendation} conduct an empirical study on where and why machine learning-based automated recommendations for method names do work or do not work, and \citet{chen2019literature} provide a more comprehensive survey that includes embeddings based on different granularities of source code such as tokens, functions or methods, sequences or method calls, binary code, and other for source code embeddings.

\section{Threats to validity}
\label{sec:threats}

We have performed a limited exploratory analysis on the ten most frequent methods in the dataset. Therefore, the results should be interpreted in the confinement of the limits of our experiment.
The results of \SVMHC depend on the features that we have extracted. Despite our best effort, it is possible that the handcrafted features can be further improved. 
Moreover, we only analyzed the ten most frequent method names. Therefore, our methodology may not generalize on different methods unless we include discriminant features for them. It is possible that experiments on other methods may produce different results.

\section{Discussions and Conclusion}
\label{sec:discussion}
The code2vec embeddings are highly-dimensional and are the results of training over millions of lines of code.
Therefore, it is nontrivial to identify the impacts, if any, of each dimension in storing semantic or syntactic characteristics of a program.
Although we really did not understand the actual meaning of each dimension of the code2vec source code embeddings, our results suggest that few handcrafted features could perform very similar to the highly-dimensional code2vec embeddings in our experiments. 
Compare to the handcrafted features, the information gains are more evenly distributed in the code2vec embeddings. Moreover, the code2vec embeddings are more resilient to the removal of dimensions with low information gains than the handcrafted features.

In this work, we described our preliminary study to demystify the source code embeddings through a comparison of the code2vec embeddings with the handcrafted features.
Although preliminary, this work provides some insights into how the features contribute to the classification task at hand.
We hope that this paper helps us to design a practical framework to objectively analyze and evaluate dimensions in the source code embeddings.
Our source code to extract \TopTen handcrafted features and train $SVM^{light}$ models for method name classification is available at \url{https://github.com/mdrafiqulrabin/handcrafted-embeddings}.

\balance
\bibliographystyle{ACM-Reference-Format}
\bibliography{refs}
\end{document}